\documentclass[a4paper,10pt]{article}

\usepackage[utf8]{inputenc}
\usepackage{fullpage}
\usepackage{amsmath}
\usepackage{amsfonts}
\usepackage{amssymb}
\usepackage{graphicx}
\usepackage{algorithmic}
\usepackage{algorithm}
\usepackage{color}
\usepackage[numbers,sort&compress]{natbib}
\usepackage{authblk}
\usepackage{subfigure}
\usepackage{multirow}
\usepackage{booktabs}
\usepackage{tikz}
\usepackage[raggedright]{sidecap}

\sidecaptionvpos{figure}{c} % position of the text in sidecaption. c=center, t=top, b=bottom

\newcommand{\argmin}{\operatornamewithlimits{arg\ min}}

\def\x{{\mathbf x}}
\def\y{{\mathbf y}}
\def\h{{\mathbf h}}

\def\K{{\mathbf K}}

\DeclareMathOperator{\var}{Var}
\DeclareMathOperator{\diag}{diag}
\def\E{{\mathbf E}}
\def\X{{\mathbf X}}
\def\Y{{\mathbf Y}}

\def\mLambda{{\boldsymbol \Lambda}}
\def\mSigma{{\boldsymbol \Sigma}}

\definecolor{darkgreen}{rgb}{0,0.6,0.2}

\title{Training Echo State Networks with Regularization through Dimensionality Reduction}

\author{Sigurd L\o{}kse \thanks{sigurd.lokse@uit.no}\thanks{Corresponding author}}
\author{Filippo Maria Bianchi\thanks{filippo.m.bianchi@uit.no}}
\author{Robert Jenssen \thanks{robert.jenssen@uit.no}}
\affil{Machine Learning Group, Department of Physics and Technology, University of Troms\o{} - The Arctic University of Norway}

\providecommand{\keywords}[1]{\textbf{\textit{Keywords---}} #1}
\begin{document}

\maketitle

\begin{abstract}
In this paper we introduce a new framework to train an Echo State Network to predict
  real valued time-series.
The method consists in projecting the output of the internal layer of the network
  on a space with lower dimensionality, before training the output layer to learn the
  target task. Notably, we enforce a regularization constraint that leads to better
  generalization capabilities.
We evaluate the performances of our approach on several benchmark tests, using different
  techniques to train the readout of the network, achieving superior
  predictive performance when using the proposed framework.
Finally, we provide an insight on the effectiveness of the implemented mechanics
  through a visualization of the trajectory in the phase space and relying on the
  methodologies of nonlinear time-series analysis.
By applying our method on well known chaotic systems, we provide evidence that the lower
  dimensional embedding retains the dynamical properties of the underlying system
  better than the full-dimensional internal states of the network.
\end{abstract}


%%%%%%%%%%%%%%%%%%%%%%%%%%%%%%%%%%%%%%%%%%%%%%%%%%%%%%%%%%%
%%%%%%%%%%%%%%%%%%%%% 1. INTRODUCTION %%%%%%%%%%%%%%%%%%%%%
%%%%%%%%%%%%%%%%%%%%%%%%%%%%%%%%%%%%%%%%%%%%%%%%%%%%%%%%%%%
\section{Introduction}
\label{sec:introduction}

% general intro
Echo State Networks (ESN) belong to the class of computational dynamical systems,
  implemented according to the so-called reservoir computing approach
  \cite{lukovsevivcius2009reservoir}.
An input signal is fed to a large, recurrent and randomly connected dynamic hidden
  layer, the \emph{reservoir}, whose outputs are combined by a memory-less layer
  called \emph{readout} to solve a specified task.
Contrary to most hard computing approaches, which demand long training procedures
  to learn model parameters through an optimization algorithm \cite{huang2005particle},
  ESN is characterized by a very fast learning procedure that usually consists in
  solving a convex optimization problem.

% works with ESN
ESN have been adopted in a variety of different contexts, such as static classification
  \cite{alexandre2009benchmarking}, speech recognition \cite{skowronski2007automatic},
  intrusion detection \cite{1504645}, adaptive control \cite{6480841} harmonic
  distortion measurements \cite{4712533} and, in general, for modeling of various
  kinds of non-linear dynamical systems \cite{han2014fuzzy}.
The application field where ESN has been used the most, is the problem of predicting
  real valued time-series relative, for example, to telephonic or electric load, where
  the forecast is usually performed 1-hour and a 24-hours ahead
  \cite{deihimi2012application, deihimi2013short, 7286732, varshneyhalf, deihimi2013short, peng2014novel}.
Outstanding results have also been achieved by ESN in prediction of chaotic time-series
  \cite{li2012chaotic, jaeger2004harnessing}, which highlighted the capability of
  these neural networks to learn amazingly accurate models to forecast a chaotic process
  from almost noise-free training data.

% Motivation
Although a large reservoir could capture the dynamics of the underlying system more accurately,
  it results in a model of increased complexity, with an inherent risk of overfitting that leads
  to lower generalization capabilities.
Additionally, several regression methods adopted to train the readout layer could be affected by
  the curse of dimensionality in case of high dimensional data, which could also cause increments in both the computational requirements in software and the resource needed in hardware \cite{bianchi2015prediction}.
Several tasks in signal processing and machine learning applications have been tackled
  by evaluating regression functions \cite{zhou2009compressed}, performing classification
  \cite{davenport2007smashed} or finding neighbors \cite{indyk1998approximate}
  in a reduced dimensional space.
In fact, in many cases it is possible to maintain meaningful distance relationships
  between original data and to deal with the curse of dimensionality at the same time.
Through dimensionality reduction, redundant features are removed, noise can be filtered
  and algorithms that are unfit for a large number of dimensions become applicable.
In the ESN literature, different methods have been proposed to increase the generalization
  ability of the network and to regularize the output.
For example, in \cite{Dutoit20091534}, the authors propose a form of regularization by shrinking
  the weights of the connections from the reservoir to the readout layer.
In \cite{Scardapane2015} by pruning some connections from the reservoir to the readout layer,
  better generalization capabilities are achieved along with some insight on which neurons are
  actually useful for the output, providing clues on how to create a good reservoir.

% contribution
In this paper we propose a novel framework for training an ESN, where an additional
  computational block is introduced to process the output of the internal reservoir,
  before being fed into the readout layer.
In particular, the internal state of the network is mapped to a properly chosen
  lower dimensional subspace, using both linear and non-linear transformations.
Accordingly, we are able to use a large reservoir to capture the dynamics of the
  underlying system, while increasing the generalization capabilities of the model
  due to implicit regularization constraints provided by dimensionality reduction
  in the recurrent layer.
Even if additional operations are introduced to compute the reduced dimensionality embedding,
  training the readout layer becomes less demanding, especially in regression 
  methods whose computational complexity depends on input dimension \cite{fodor2002survey}.
  %(I don't know if this is really that relevant. Remember that the nu-SVR only depends on the kernel function [inner products]. Although one has to calculate distances to compute the kernel function...)
With the proposed procedure we improve the generalization capabilities of the network, achieving better results on well-known benchmarking problems
  with respect to the standard ESN architecture.
Additionally, in cases where data can be mapped to spaces with 2 or 3 dimensions, 
  internal network dynamics can be visualized precisely and relevant patterns can be detected.
To justify the results obtained and to understand the mechanisms which determines
  the effectiveness of the proposed system, we provide a theoretical study based
  on methods coming from the field of nonlinear time-series analysis.
To the best of the authors' knowledge, the coupling of dimensionality reduction with the
  ESN architecture has not been explored before.

% paper organization
The remainder of the paper is organized as follows.
In Sect. \ref{sec:background} we describe the ESN structure along with existing
  approaches for its training and we review the dimensionality reduction methods
  adopted in this work.
In Sect. \ref{sec:architecture} we present our proposed architecture, providing
  implementation details.
In Sect. \ref{sec:experiments} we describe the datasets used to test our system,
  the experimental settings adopted and the performance reached on several prediction problems.
In Sect. \ref{sec:discussion} we analyze the results and the functioning of our
  system through the perspective of nonlinear time-series analysis.
Finally, in Sect. \ref{sec:conclusion} we draw our conclusions.

%%%%%%%%%%%%%%%%%%%%%%%%%%%%%%%%%%%%%%%%%%%%%%%%%%%%%%%%%
%%%%%%%%%%%%%%%%%%%%% 2. BACKGROUND %%%%%%%%%%%%%%%%%%%%%
%%%%%%%%%%%%%%%%%%%%%%%%%%%%%%%%%%%%%%%%%%%%%%%%%%%%%%%%
\section{Background material}
\label{sec:background}

In the following, we shortly review the methodologies adopted in our framework.
Initially, we describe the classic ESN architecture and two effective approaches
  adopted for its training.
Successively, we summarize two well-know methods used for reducing the
  dimensionality of the data and for mapping them in a smaller subspace.

% ------------- ESN -------------
\subsection{Echo state Network}
\label{sec:esn}

\begin{SCfigure}[1.2][h!]
    \centering
    \includegraphics[scale=0.8, keepaspectratio]{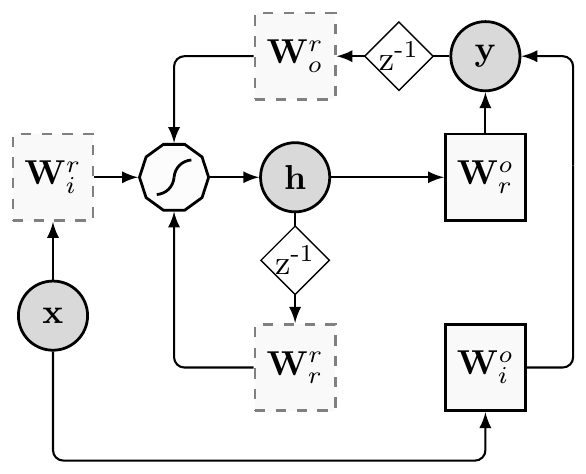}
    \caption{\footnotesize Schematic depiction of the ESN architecture.
    The circles represent input $\mathbf{x}$, state, $\mathbf{h}$, and output, $\mathbf{y}$, respectively.
    Solid squares $\mathbf{W}_{r}^{o}$ and $\mathbf{W}_{i}^{o}$, are the trainable matrices, respectively,
     of the readout, while dashed squares, $\mathbf{W}_{r}^{r}$, $\mathbf{W}_{o}^{r}$, and $\mathbf{W}_{i}^{r}$,
     are randomly initialized matrices.
    The polygon represents the non-linear transformation performed by neurons and $\text{z}^{\text{-1}}$ is the unit delay operator.}
    \label{fig:esn}
\end{SCfigure}

% architecture description
An ESN consists of a large, untrained recurrent layer of non-linear units and a
  linear, memory-less readout layer, usually trained with a linear regression.
A visual representation of an ESN is reported in Fig. \ref{fig:esn}

The equations describing the ESN state-update and output are, respectively,
  defined as follows:
\begin{align}
\label{eq:esn_state_update}
\mathbf{h}[k] =& \phi(\mathbf{W}_{r}^{r}\mathbf{h}[k-1] + \mathbf{W}_{i}^{r}\mathbf{x}[k] + \mathbf{W}_{o}^{r}\mathbf{y}[k-1] + \xi),\\
\label{eq:esn_esn_output}
\mathbf{y}[k] =& \mathbf{W}_{i}^{o}\mathbf{x}[k] + \mathbf{W}_{r}^{o}\mathbf{h}[k],
\end{align}
  where $\xi$ is a small i.i.d. noise term.
The reservoir consists of $N_r$ neurons characterized by a transfer/activation
  function
  $\phi(\cdot)$,
  typically implemented as a hyperbolic tangent function.
At time instant $k$, the network is driven by the input signal
  $\mathbf{x}[k]\in \mathbb{R}^{N_i}$
  and generates the output
  $\mathbf{y}[k] \in \mathbb{R}^{N_o}$,
  being $N_i$ and $N_o$ the dimensionality of input and output, respectively.
The vector $\mathbf{h}[k] \in \mathbb{R}^{N_r}$ describes the ESN (instantaneous) state.
The weight matrices
  $\mathbf{W}_r^r \in \mathbb{R}^{N_r \times N_r}$
  (reservoir connections),
  $\mathbf{W}_i^r \in \mathbb{R}^{N_r \times N_i}$
  (input-to-reservoir), and
  $\mathbf{W}_o^r \in \mathbb{R}^{N_r \times N_o}$
  (output-to-reservoir feedback) contain real values in the $[-1, 1]$ interval,
  sampled from a uniform distribution.

According to the ESN theory, the reservoir $\mathbf{W}_r^r$ must satisfies the
  so-called ``echo state property'' (ESP) \cite{lukovsevivcius2009reservoir}.
This guarantees that the effect of a given input on the state of the reservoir
  vanish in a finite number of time intervals.
A widely used rule-of-thumb suggests to rescale the matrix
  $\mathbf{W}_r^r$
  to have
  $\rho(\mathbf{W}_r^r) < 1$,
  where $\rho(\cdot)$ denotes the spectral radius, but several theoretically-founded
  approaches have been proposed in the literature to properly tune $\rho$ in an ESN driven by a specific input
  \cite{boedecker2012information, bianchi2016investigating, verstraeten2009quantification}.

The weight matrices $\mathbf{W}_{i}^{o}$ and $\mathbf{W}_{r}^{o}$ instead, are
  optimized for the task at hand.
To determine them, let us consider the training sequence sequence of $T_{tr}$
  desired input-outputs pairs given by:
\begin{equation}
(\mathbf{x}[1], y^*[1]) \ldots, (\mathbf{x}[T_{tr}], y[T_{tr}]),
\end{equation}

In the initial phase of training, called \emph{state harvesting}, the inputs are
  fed to the reservoir in accordance with Eq.~\ref{eq:esn_state_update},
  producing a sequence of internal states
  $\mathbf{h}[1], \ldots, \mathbf{h}[T_{tr}]$.
Since, by definition, the outputs of the ESN are not available for feedback,
  according to the \emph{teacher forcing} procedure, the desired output is used
  instead in Eq.~\ref{eq:esn_esn_output}.
States are stacked in a matrix
  $\mathbf{S} \in \mathbb{R}^{T_{tr} \times N_i + N_r}$
  and the desired outputs in a vector $\mathbf{y}^* \in \mathbb{R}^T_{tr}$:
\begin{align*}
\mathbf{S} = & 
\left[\begin{array}{c}
\mathbf{x}^T[1], \,\, \mathbf{h}^T[1] \\
\vdots \\
\mathbf{x}^T[T_{tr}], \,\, \mathbf{h}^T[T_{tr}]
\end{array}\right], \,
\mathbf{y}^* =
\left[\begin{array}{c}
y^*[1] \\
\vdots \\
y^*[T_{tr}]
\end{array}\right].
\end{align*}
The initial $D$ rows $\mathbf{S}$ and $\mathbf{y}^*$ are the washout elements
  that should be discarded, since they refer to a transient phase in the ESN's
  behavior.

Since the gain of the sigmoid non-linearity in the neurons is largest around
  the origin, three coefficients $\omega_i$, $\omega_o$ and $\omega_f$ are used
  to scale the input, desired output and feedback signals respectively.
In this way, it is possible to control the amount of non-linearity introduced
  by the processing units.

The training of the readout consists in solving a convex optimization problem,
  for which several closed form solution have been proposed in the literature.
The standard procedure to train the readout, originally proposed in
  \cite{jaeger2001echo}, consists in a regularized least-square regression,
  which can be easily computed through the Moore-Penrose pseudo-inverse.
However, to learn the optimal readout we also consider the Support Vector Regression (SVR),
a supervised learning model that can efficiently perform a non-linear separation
  of data using a kernel function to map the inputs into high-dimensional feature
  spaces, where they are linearly separable \cite{burges1998tutorial}.

% Readout Training ------ Ridge Regression
\paragraph{Ridge Regression:}
to train the readout with a linear regressor we adopted ridge regression, whose
  solution can be computed by solving the following regularized least-square problem:
\begin{equation}
\mathbf{W}_{\text{ls}}^* = \argmin_{\mathbf{W} \in \mathbb{R}^{N_i+N_r}}  \frac{1}{2} \| \mathbf{S}\mathbf{W} - \mathbf{y}^* \|^2 + \frac{\lambda}{2} \|\mathbf{W}\|^2 = \left( \mathbf{S}^T\mathbf{S} + \lambda \mathbf{I} \right)^{-1}\mathbf{S}^T \mathbf{y}^*\,,
\label{eq:esn_opt}
\end{equation}
where $\mathbf{W} = \left[ \mathbf{W}_i^o \, \mathbf{W}_r^o \right]^T$ and $\lambda \in \mathbb{R}^+$ is the $\mathrm{L}_2$ regularization coefficient. 

% Readout Training ------ Support Vector Regression
\paragraph{Support Vector Regression:} 
we adopt a $\nu$-SVR \cite{scholkopf2000new} with a Gaussian kernel, initially
  proposed in \cite{bianchi2015prediction} as method for readout training.
In this case, the ESN acts as a preprocessor for a  $\nu$-SVR kernel and their
  combination can be seen as an adaptive kernel, capable of learning a task-specific
  time dependency.
The state
  $\mathbf{s}_i = \left[ \mathbf{x}^T[i] \,\, \mathbf{h}^T[i] \right]^T$
  is projected to a higher dimensional feature space
  $\phi(\mathbf{s}_i)$,
  and the $\nu$-SVR is applied on the resulting space.
The dual optimization problem can be written as:
\newcommand{\alphadiff}{\ensuremath{\left(\boldsymbol{\alpha} - \boldsymbol{\alpha}^*\right)}}
\begin{equation}
\mathbf{W}_{\text{svr}}^* = 
\left\{%
\begin{aligned}
\underset{\mathbf{\alpha},\mathbf{\alpha}^* \in \mathbb{R}^{T_{tr}}}{\text{min}}
& \;\; \frac{1}{2}\alphadiff\mathbf{K}\alphadiff + \mathbf{y}^{*T}\alphadiff \\
\text{subject to}
& \;\; \mathbf{1}^T\alphadiff = 0 \,, \\
& \;\; \mathbf{1}^T \left(\boldsymbol{\alpha} + \boldsymbol{\alpha}^*\right) \le \lambda\nu \,, \\
& \;\; 0 \le \alpha_i, \alpha_i^* \le \dfrac{\lambda}{T_{tr}}, \, i = \, \ldots, T_{tr}
\end{aligned}\right.
\label{eq:esn_opt_svr_dual}
\end{equation}
  where each entry $K_{ij}$ is given by
  $\mathcal{K}\left(\mathbf{s}_i, \mathbf{s}_j \right)$,
  with
  $\mathcal{K}(\cdot, \cdot)$
  being a reproducing Gaussian kernel associated to the feature mapping, given by
  $\mathcal{K}(\mathbf{s}_i, \mathbf{s}_j) = \exp\left\{ - \gamma \|\mathbf{s}_i - \mathbf{s}_j\|^2 \right\}$,
  where $\gamma$ is denoted as the scale parameter.

By an extension of the representer's theorem, the output of the ESN at a generic
  time-instant $t$ in this case is given by:
\begin{eqnarray}
y[t] = \sum_{i=1}^{T_{tr}} \left( \alpha_i - \alpha_i^* \right)\mathcal{K}\left(\mathbf{s}_i, \mathbf{s}_t\right) \,,
\label{eq:esn_weights_dual}
\end{eqnarray}
  where $\alpha_i$ and $\alpha_i^*$ are the entries of the optimal solution to
  problem Eq.~\ref{eq:esn_opt_svr_dual}, and they are non-zero only for patterns
  that are support vectors.

% --------------- DIMENSIONALITY REDUCTION METHODS ---------------
\subsection{Dimensionality reduction methods}
\label{sec:dimred}

In the following, we describe the dimensionality reduction techniques that we
  implemented in our framework.
First of all, we underline that several approaches can be followed for reducing
  the dimensionality of the data and to learn underlying manifold on a subspace
  of the data space \cite{van2009dimensionality, 4912217, 6530774}.
In this work, we limit our analysis to the well know and effective, yet simple procedures, namely
  Principal Component Analysis (PCA) \cite{Hotelling1933} and kernel Principal Component Analysis
  (kPCA) \cite{scholkopf1997kernel}.
% Many other options are available and could be considered in a future work. - repetition

% PCA
\paragraph{PCA} is a statistically motivated method, which projects  the data onto
  an orthonormal basis that preserves most variance in the input signal, while
  ensuring that the individual components are uncorrelated. These basis vectors are
  called the \textit{Principal Components}.
Let $\X \in \mathbb{R}^p$ be a random vector and let $\mSigma_X = \E \mLambda \E^T$
  be its covariance matrix, where
  $\E = \begin{pmatrix} \mathbf e_1 & \mathbf e_2 & \cdots & \mathbf e_p \end{pmatrix}$
  and
  $\mLambda = \diag(\lambda_i)$
  is the orthogonal eigenvector matrix and the diagonal eigenvalue matrix respectively.
Then the linear transformation $\Y = \E^T \X$ ensures that the covariance matrix of
  $\Y$ is $\mSigma_Y = \mLambda$,
  which clearly implies that the components of $\Y$ are uncorrelated.
We also see that
\begin{equation}
\label{eq:pca_variance}
  \sum_{i = 1}^p \var X_i = \sum_{i = 1}^p \var Y_i = \sum_{i = 1}^p \lambda_i.
\end{equation}

To reduce the dimensionality to $d$ dimensions, we project the data onto the $d$
  eigenvectors with the largest eigenvalues.
That is,
\begin{equation*}
  \widehat\Y = \E_d^T \X,
\end{equation*}
  where
  $\E_d = \begin{pmatrix}\mathbf e_1 & \mathbf e_2 & \cdots & \mathbf e_d\end{pmatrix}$
  is the \textit{truncated} eigenvector matrix associated with the eigenvalues
  $\lambda_1 \geq \lambda_2 \geq \cdots \geq \lambda_d$.
According to Eq.~\ref{eq:pca_variance}, this ensures that $\widehat\Y$ preserves
  most of the variance of $\X$.

% ------------ kPCA ------------
\paragraph{Kernel Principal Component Analysis} (kPCA) is a nonlinear extension of PCA.
Given a valid positive semidefinite (psd) Mercer Kernel
	\[
    \mathcal{K}(\h[i], \h[j]) = \langle \Phi(\h[i]), \Phi(\h[j]) \rangle_\mathcal{H},
	\]
	where $\Phi$ is some nonlinear mapping from feature space to a Hilbert space $\mathcal H$,
	Kernel PCA implicitely performs PCA in $\mathcal H$.

Let $\K = \{K_{ij}\}_{N\times N}$, where $K_{ij} = \mathcal{K}(\h[i], \h[j])$ be the \textit{kernel matrix}
	and let $\E$ and $\mLambda$ be its eigenvector and eigenvalue matrix respectively with the eigenvalues
	sorted in descending order.
Then the projection of the in-sample data onto the principal components in $\mathcal H$ is given by
\begin{equation} \label{eq:is_kpca}
  \mathbf{\bar{H}} = \E\mLambda^{\frac{1}{2}}.
\end{equation}
The out-of-sample approximation for the projection of a data point $\h[k]$ onto the
	$\ell$th principal component is given by
\begin{equation} \label{eq:oos_kpca}
  \bar{h}_\ell[k] = \frac{1}{\sqrt{\lambda_\ell}} \sum_{i = 1}^N e_\ell(i) \mathcal{K}(\h[i], \h[k]).
\end{equation}
Just like canonical PCA, to perform dimensionality reduction with kPCA, one need to use the truncated
	eigenvector- and eigenvalue matrix with Eq.~\ref{eq:is_kpca} and Eq.~\ref{eq:oos_kpca}.

The kernel function that is commonly used in practice is the \textit{Gaussian kernel} which is given by
  $\mathcal{K}(\h[i], \h[j]) = \exp\left\{-\gamma \|\h[i] - \h[j]\|^2 \right\}$, where $\gamma$ controls
	the width of the kernel.

Both PCA and kPCA methods admit an out of sample extension, a feature which is required in our framework, as discussed later.

%%%%%%%%%%%%%%%%%%%%%%%%%%%%%%%%%%%%%%%%%%%%%%%%%%%%%%%%%%%%%%%%
%%%%%%%%%%%%%%%%%%% 3. PROPOSED ARCHITECTURE %%%%%%%%%%%%%%%%%%%
%%%%%%%%%%%%%%%%%%%%%%%%%%%%%%%%%%%%%%%%%%%%%%%%%%%%%%%%%%%%%%%%
\section{Proposed architecture}
\label{sec:architecture}

In this section, we provide the details of the architecture of the framework proposed.

The large size of the reservoir, specified by the amount $N_r$ of hidden neurons,
  is one of the main features that determines the effectiveness of the reservoir
  computing paradigm. Due to the high quantity of neurons, the internal recurrent
  connections in the reservoir are capable of generating a rich and heterogeneous
  dynamic to solve complex memory-dependent tasks.
However, as the size of the reservoir increases, also the complexity of the model
  grows, with a consequent risk of overfitting caused by a reduced generalization
  capability \cite{belkin2006manifold}.
Dimensionality reduction and manifold learning are techniques that allows to
  diminish the variance in the data and to introduce a bias, which can reduce the
  expected value on the prediction error \cite{friedman1997bias}.
In the architecture proposed, we use a large reservoir in order to capture the
  dynamic of the underlying unknown process and then, through a dimensionality
  reduction procedure, we enforce regularization constraints to increase the
  generalization capability of our model.
Another important consequence that follows from reducing the dimensionality of
  the reservoir is that complex regression methods can benefit from a reduced
  computational complexity if the internal states are described by a lower number of variables.
Additionally, several methods used to identify, in an unsupervised way, the
  configurations of hyperparameters which maximize the computational capabilities
  of the network, require computational demanding procedures of analysis
  \cite{bianchi2016investigating, livi2016determination, boedecker2012information}.
These procedures would greatly benefit from the simplification offered by our
  proposed architecture.

At each time step $t$, the vector $\h[t] \in \mathbb{R}^{N_r}$ that represents the
  internal state of the reservoir, is mapped into a lower dimensional space by a
  projector
  $\mathcal{P}: \mathbb{R}^{N_r} \rightarrow \mathbb{R}^{d}$.
The new $d$-dimensional state vector
  $\bar{\h}[t] = \mathcal{P}(\h[t])$
  is then processed by the readout to compute the predicted value $\y[t]$.

To train our system, the time-series is split in three contiguous parts, namely
  the training
  $\{ \mathbf{X}_{tr}, \mathbf{Y}_{tr} \}$,
  validation
  $\{ \mathbf{X}_{vs}, \mathbf{Y}_{vs} \}$
  and test set
  $\{ \mathbf{X}_{ts}, \mathbf{Y}_{ts} \}$.
Since we deal with time-series prediction problems, each set contains coupled
  real values, which represent the input value and the ground truth of the
  associated prediction.
For example in the training set we have
  $\{\x[t], \y[t] \}_{t=1}^{T_{tr}}$,
  where $\y[t]$ is the predicted value of $\x[t]$.
The regression function in the readout is implemented according to one of the
  two procedures proposed in Sect. \ref{sec:esn} and the model parameters are
  learned on the training data.
The system depends on several hyperparameters, which affects the network
  behavior and they must be carefully tuned on the specific problem at hand by
  performing a cross-validation procedure on the validation set, with a method
  whose details are provided in the next section.

Once the model has been trained, a new test element $\x[t]$ of the test set is
  processed and the relative internal reservoir state $\h[t]$ is generated.
Successively, the projection $\bar{\h}[t]$ in the subspace with reduced dimensionality
  is evaluated using a suitable out of sample approximation.
In the case of PCA this can be done by projecting $\h[t]$ on the basis defined by
  the covariance matrix computed on the $T_{tr}$ states relative to the elements
  in training set, which are collected in the matrix $\mathbf{H}_{tr}$ during the
  training phase.
For kPCA it is possible to use the N\"{y}strom approximation \cite{baker1973numerical},
  which specifies an interpolating function for determining the values of out of samples
  data points. 

\begin{figure}[ht]
	\centering
	\includegraphics[width=0.9\textwidth]{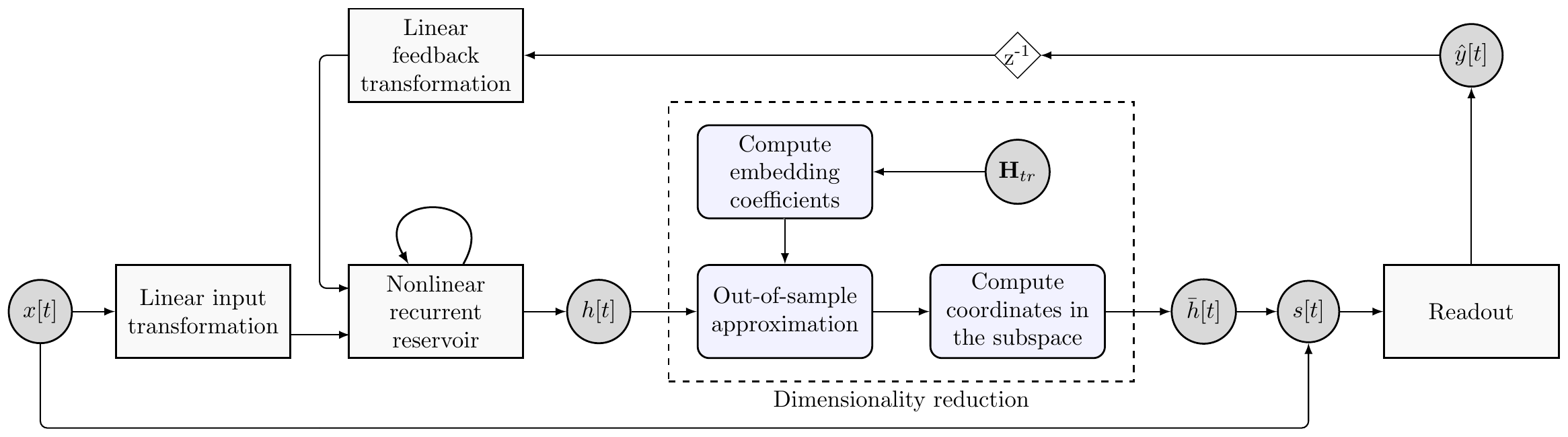}
	\caption{\footnotesize When a new element $\x[t]$ is fed into the network, the internal state of the ESN is updated and its new value is stored in $\h[t]$. Such state vector is then projected on a subspace, computed during the training on the state matrix $\mathbf{H}_{tr}$ and the vector of reduced dimensionality in this subspace $\bar{\h}[t]$ is evaluated. At this point, the predicted output value $\hat{\y}[t]$ is computed by the ESN readout.}
	\label{fig:oos}
\end{figure}

A schematic representation of the whole procedure is depicted in Fig. \ref{fig:oos}.

% ----------- Hyperparameter optimization -----------
\subsection{Hyperparameter optimization}
\label{sec:hyper}

The set of hyperparameters $\boldsymbol{\theta}$ that are used to control the
  architecture of the ESN, the regression in the readout and the dimensionality-reduction
  procedure are optimized by minimizing a loss function $L( \cdot)$, defined as
\begin{equation}
\label{eq:loss}
L( \boldsymbol{\theta}_i ) = (1 - \alpha)Err(\mathbf{Y}_{vs}) + \alpha \theta_i^{\text{(d)}},
\end{equation}
  where $\theta^{\text{(d)}} = \frac{d}{N_r}$ is the hyperparameter that defines
  the number of dimensions, $d$, of the new subspace.
In order to lower the complexity of the model, $L( \cdot)$ jointly penalizes
  prediction error on the validation set and the number of dimensions retained
  after the dimensionality reduction.

\begin{figure}[ht]
	\centering
	\includegraphics[width=0.9\textwidth]{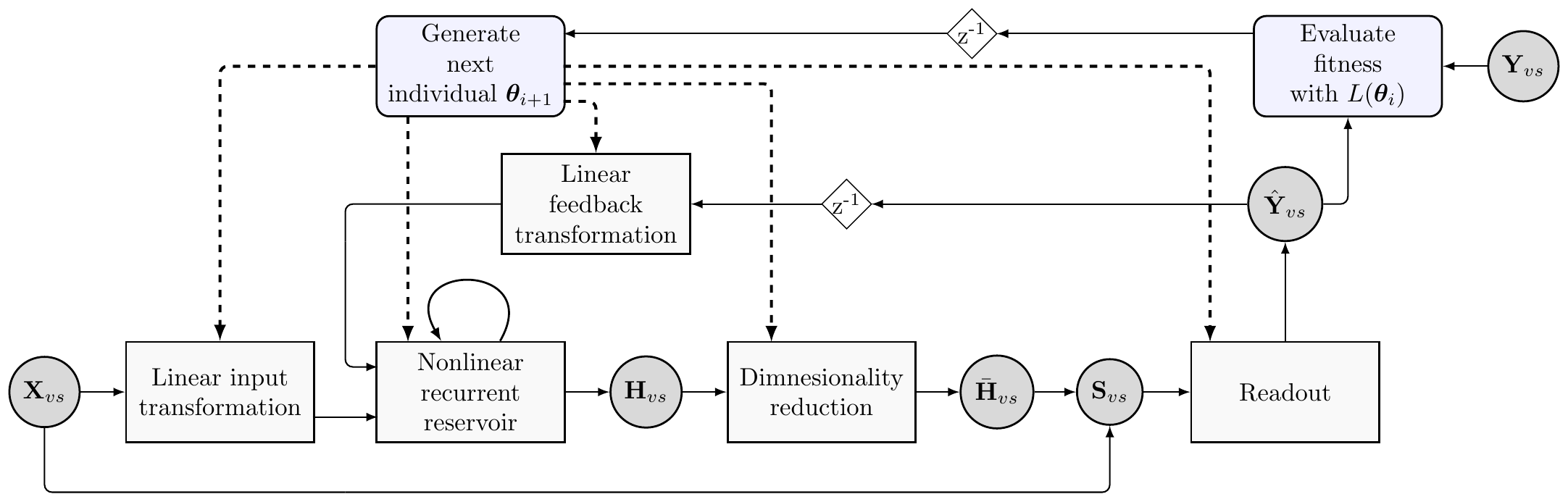}
	\caption{\footnotesize Overview of the hyperparameters optimization in the proposed architecture. At the $i$-th iteration, the input elements of the validation set $\mathbf{X}_{vs}$ are processed by the ESN configured with the hyperparemeters in $\boldsymbol{\theta}_i$, which is the $i$-th individual generated by the GA. The predicted output $\hat{\mathbf{Y}}_{vs}$ produced by the network is matched against the ground truth $\mathbf{Y}_{vs}$, the resulting similarity (prediction error) is used to compute the fitness of $\boldsymbol{\theta}_i$ with the loss function $L( \boldsymbol{\theta}_i )$. In the next iteration, a new individual $\boldsymbol{\theta}_{i+1}$ is generated, depending on results obtained so far and on the policies of the GA.}
	\label{fig:architecture}
\end{figure}

The loss function is minimized using a standard genetic algorithm with Gaussian mutation,
  random crossover, elitism and tournament selection \cite{294849}.
While the hyperparameter optimization is performed on the validation set,
  the best individual found is stored and it is successively used to configure the network during the training phase.
A schematic description of the training procedure is depicted in Fig. \ref{fig:architecture}.

\section{Experiments}
\label{sec:experiments}

% experimental setup description
The component of the loss function (Eq. \ref{eq:loss}) relative to the error on the given task, is implemented by the Normalized Root
  Mean Squared Error (NRMSE):
\begin{equation*}
\label{eq:nrmse}
\textrm{NRMSE} = \sqrt{\frac{\langle \lVert \mathbf{y}[k] - \mathbf{y}^*[k] \rVert^2 \rangle}{\langle \lVert \mathbf{y}[k] - \langle\mathbf{y}^*[k]\rangle \rVert^2 \rangle}},
\end{equation*}
where $\mathbf{y}[k]$ is the ESN prediction and $\mathbf{y}^*[k]$ the desired/teacher output.

The GA uses a population size of 50 individuals and evaluates 20 generations.
The individuals are mutated and bred at each generation with a mutation probability
  of $P_\text{mut} = 0.2$ and a crossover probability of $P_\text{cx} = 0.5$.
The individuals in the next generation are selected by a tournament strategy with
  a tournament size of 4 individuals.
The bounds for all parameters are shown in Tab.~\ref{tab:genopt_parameters}.
The weight parameter $\alpha$ in the loss function (Eq.~\ref{eq:loss}) is set to 0.1.

Due to the stochastic nature of the ESN, which is a consequence of the random initialization
  of the weight matrices $\mathbf{W}_i^r$, $\mathbf{W}_r^r$ and $\mathbf{W}_o^r$, 
  each individual is evaluated on the validation
  set using 5 networks initialized with different weight parameters.
The fitness is then given by the NRMSE, averaged over these 5 networks.
Once the optimal set of parameters $\boldsymbol\theta^*$ has been found, we predict
  values for the test set using 32 randomly initialized networks, using the same
  set of optimal parameters.
\bgroup
\def\arraystretch{1.3} %vertical padding
\setlength\tabcolsep{.4em} %horizontal padding
\begin{center}
\begin{table}[tbp]\scriptsize
\label{tab:genopt_parameters}
\centering
\caption{\footnotesize Each hyperparameter is searched by the GA in the interval [\emph{min}, \emph{max}] with resolution $\sigma$. 
  The fields in the table are the following: spectral radius of the reservoir ($\rho$),
  neurons in the reservoir ($N_r$),
  noise in ESN state update ($\xi$),
  scaling of input,
  teacher and feedback weights ($\omega_i$, $\omega_o$, $\omega_f$),
  embedding dimension $\left( \theta^\text{(d)} = \frac{d}{N_\text{r}} \right)$,
  $\mathrm{L}_2$ norm regularization factor ($\lambda$),
  $\nu$-SVR parameters ($C$, $\gamma$, $\nu$).}
\vspace{0.2cm}
\begin{tabular}{r|ccccccccccc}

\cmidrule[1.5pt]{1-12}
  \multicolumn{1}{c}{}              &
  $\mathbf{N_\text{r}}$             &
  $\boldsymbol{\xi}$                &
  $\boldsymbol{\omega_i}$           &
  $\boldsymbol{\omega_o}$           &
  $\boldsymbol{\omega_f}$           &
  $\boldsymbol{\rho}$               &
  $\boldsymbol{\theta^\text{(d)}}$  &
  $\boldsymbol{\gamma}$             &
  $\boldsymbol{\lambda}$            &
  $\mathbf{C}$                      &
  $\boldsymbol{\nu}$                \\
\cmidrule[1.5pt]{1-12}
  \emph{min}  & 100 & 0.0   & 0.1   & 0.1   & 0.0   & 0.5   & 0.001 & 0.001 & 0.001 & 0.001 & 0.001 \\
  \emph{max}  & 500 & 0.1   & 0.9   & 0.9   & 0.6   & 1.4   & 1.0   & 0.1   & 1.0   & 10.0  & 1.0   \\
  $\sigma$    & 5   & 0.01  & 0.08  & 0.08  & 0.06  & 0.09  & 0.1   & 0.01  & 0.1   & 1.0   & 0.1   \\

\cmidrule[1.5pt]{1-12}

\end{tabular}
\end{table}
\end{center}
\egroup

% ---------- dataset description ----------
\subsection{Datasets description}

To test our system, we consider 3 benchmark tasks commonly used in time-series
  forecasting, namely the prediction of Mackey-Glass time-series, of multiple
  superimposed oscillator and of the NARMA signal.
The forecasting problems that we consider have a different level of difficulty,
  given by the nature of the signal and the complexity of the prediction task.
Accordingly to a commonly used approach \cite{jan2005did}, in each prediction
  task we set the forecast step $\tau_f$ by computing a statistic that measures
  the independence of $\tau_f$-separated points in the time series.
One usually wants the smallest $\tau_f$ that guarantees the measurements to be
  decorrelated.
  Hence, we considered the first zero of the autocorrelation function of the time
  series, which yields the smallest $\tau_f$ that maximizes the linear independence
  between the samples.
Alternatively, it is possible to choose the forecast step by considering more general
  forms of independence, such as the first local minimum on the average mutual
  information \cite{fraser1986independent} or on the correlation sum \cite{liebert1989proper}.

\paragraph{Mackey-Glass time-series:} the input signal is generated from the Mackey-Glass (MG) time-delay differential system, described by the following equation:
\begin{equation*}
    \label{eq:MG_signal}
    \frac{dx}{dt} = \frac{\alpha x(t-\tau_{\mathrm{MG}})}{1+ x(t-\tau_{\mathrm{MG}})^{10}} - \beta x(t).
\end{equation*}
We generated a time-series of 150000 time-steps using $\tau_{\mathrm{MG}} = 17, \alpha = 0.2, \beta = 0.1$, initial condition $x(0)=1.2$, and 0.1 as integration step for (\ref{eq:MG_signal}).

\paragraph{NARMA signal:} the Non-Linear Auto-Regressive Moving Average (NARMA) task, originally proposed in \cite{jaeger2002adaptive}, consists in modeling the output of the following $r$-order system:
\begin{equation*}
    \label{eq:narma_signal}
    y(t + 1) = 0.3y(t) + 0.05y(t)\left[\sum \limits_{i=0}^{r} y(t - i)\right] + 1.5x(t - r)x(t) + 0.1. 
\end{equation*}
The input to the system $x(t)$ is a uniform random noise in [0, 1], and the model is trained to reproduce $y(t+1)$.
The NARMA task is known to require a memory of at least $r$ past time-steps, since the output is determined by the current input and outputs relative to the last $r$ time-steps.
In our test we set $r=20$.

\paragraph{Multiple superimposed oscillator:}
The prediction of a sinusoidal signal is a relatively simple task, which demands a minimum amount of memory to determine the next network output.
However, superimposed sine waves with not integer frequencies are much harder to predict, since the wavelength can be extremely long.
The signal we consider is the multiple superimposed oscillator (MSO), studied in \cite{jaeger2004harnessing} and defined as:
\begin{equation*}
    \label{eq:mso_signal}
    y(t) = sin(0.2t) + sin(0.311t) + sin(0.42t) + sin(0.51t) + sin(0.63t) + sin(0.74t)
\end{equation*}
ESN struggles to solve this task, since neurons in the reservoir tends to couple, while the task requires the simultaneous existence of multiple decoupled internal states \cite{wierstra2005modeling}.

% ---------- Results ----------
\subsection{Results}

\begin{figure}[ht]
\centering
	\subfigure[Mackey-Glass]{
	\includegraphics[width=13.5em,height=10em]{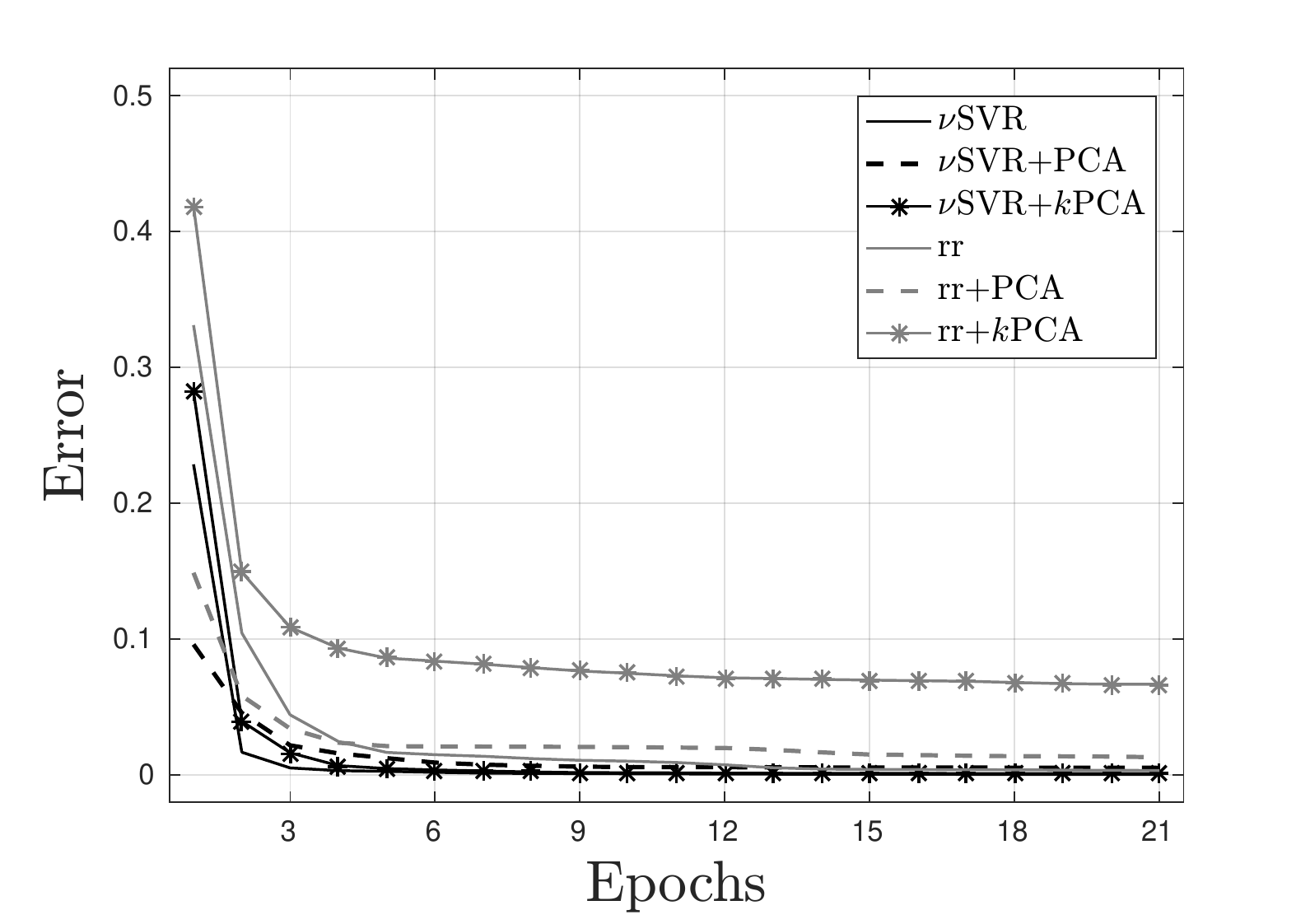}
	\label{fig:MG_train}
	}
	~
	\subfigure[NARMA]{
	\includegraphics[width=13.5em,height=10em]{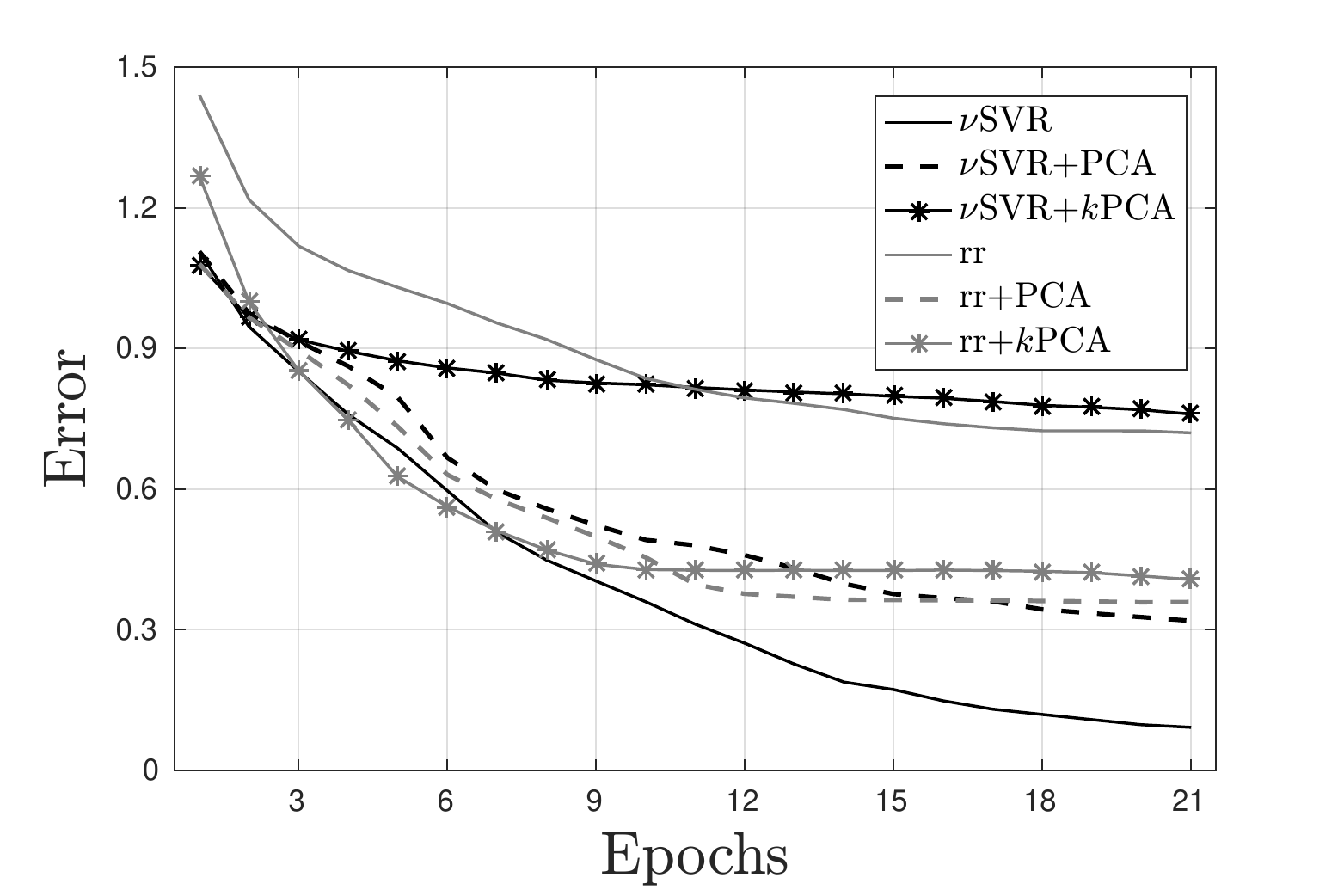}
	\label{fig:NARMA_train}
	}
	~
	\subfigure[Multiple Superimposed Oscillators]{
	\includegraphics[width=13.5em,height=10em]{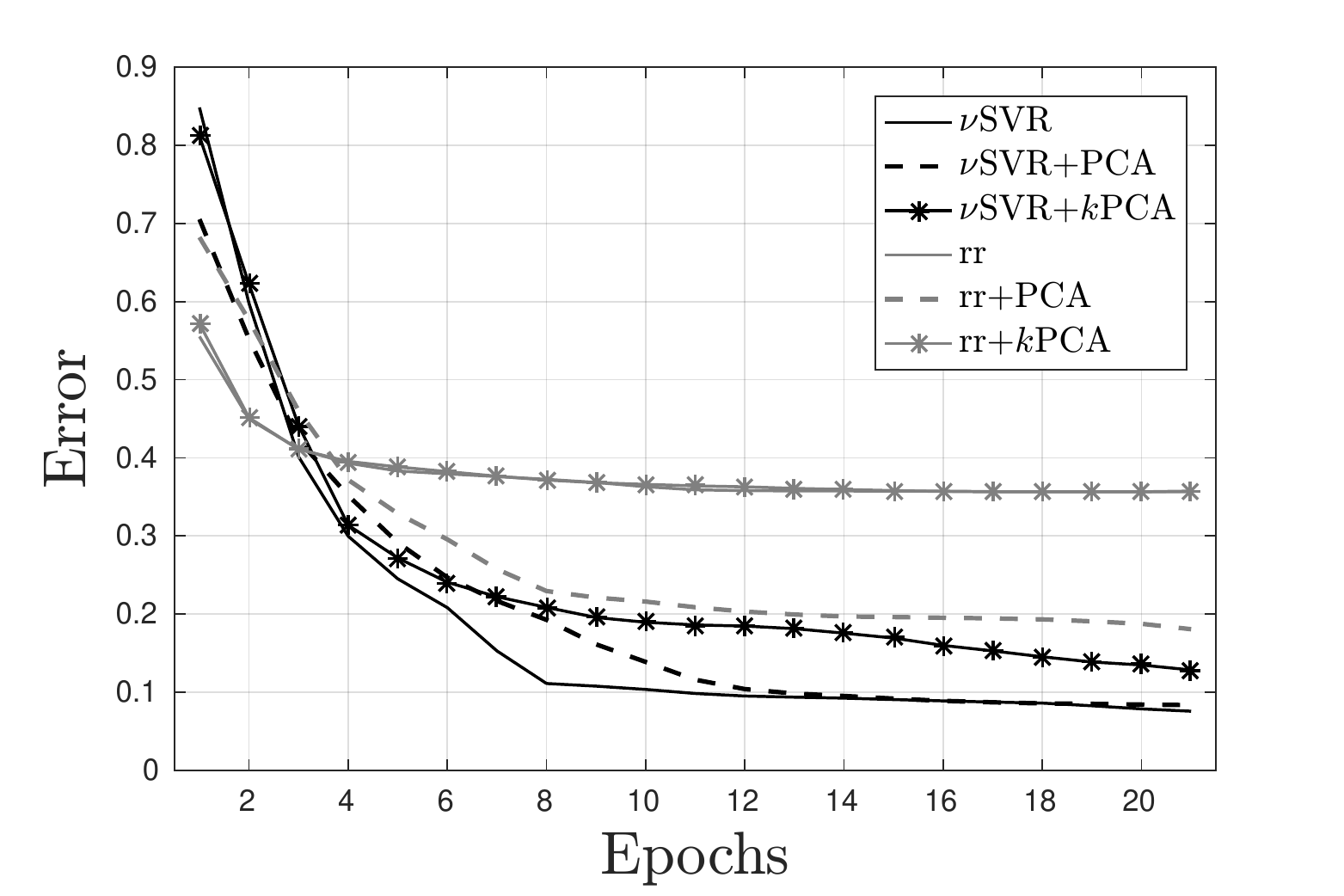}
	\label{fig:MSO_train}
	}
\caption{\footnotesize Convergence of the error on the validation set in hyperparameters optimization with the GA. Black lines represent models whose readout is trained with $\nu$-SVR, models trained with ridge regression are depicted with gray lines.}
\label{fig:train_conv}
\end{figure}

% discussion of the results
The averaged prediction results and the standard deviations are reported in Tab.~\ref{tab:results}.
The convergence rate during the optimization of the hyperparameters for each method, expressed as the NRMSE error on the validation set, is depicted in Fig. \ref{fig:train_conv}.

% MG
The prediction of MG is a quite simple task and each model manages to achieve high forecast accuracy.
However, by applying a dimensionality reduction on the states of the reservoir, it is possible to lower the error by one or more order of magnitude.
Also the standard deviation of the prediction error decreases, especially in the models using kPCA.
The best results are achieved by $\nu$-SVR + PCA and $\nu$-SVR + kPCA, while using $\nu$-SVR without reducing reducing the dimensions of the reservoir demonstrated to be less effective.
This means that non-linearities benefits the training, but without enforcing the regularization constraint the complexity of the model is to high to fit well testing points.
As we can see, in every case the number of dimensions $d$ retained by both PCA and kPCA is much lower than the optimal number of neurons $N_r$ identified.
This underline the effectiveness of the regularization conveyed by our architecture.
From Fig. \ref{fig:MG_train} results that the model implementing ridge regression + kPCA achieves the lowest convergence rate during the cross-validation step.
However, thanks to the generalization power provided by the nonlinear dimensionality reduction, the test error is lower than the other models, whose readout is trained with ridge regression.

% NARMA
In NARMA prediction task, the best result is achieved by training the readout function with $\nu$-SVR on a reservoir output, whose dimensionality is reduced by kPCA. 
NARMA is a more complex task which requires a higher amount on nonlinearity to be solved.
This is clearly reflected by the results, which improve as more nonlinearity is introduced to learn the function, both in the readout training and in the dimensionality reduction procedure.
At the same time, the bias introduced by the regularization enhance the generalization capability of the network significantly.
For what concerns the number of dimensions of the optimal subspace, it is higher than in MG task, except for the model implemented with ridge regression + PCA.
In this latter cases, however, we obtain the worst performance.
Interestingly, from Fig. \ref{fig:NARMA_train} we observe that kPCA has the lower convergence rate, even if this is the best performing model in the testing phase.
In this case, the dimensionality reduction introduces a bias, which prevents the model to overfit on the validation data and to develop a high predictive power.
On the other hand, the model with $\nu$-SVR and no dimensionality reduction, overfits on the validation data with a consequent poor performance in the test phase. 

% MSO
Finally, in the MSO task the model with the highest prediction performance is $\nu$-SVR without the dimensionality reduction.
In this case, the signal to predict has an extremely long periodicity, which demands a high amount of memory in the network.
Hence, the compression of the information through the dimensionality reduction could hamper the memory capacity of the network.
Furthermore, due to the long periodicity, the slice of time-series used to train the network can be quite different from the slice to be predicted in the test.
Consequently, test points are projected in a subspace which is not optimal, as the basis is learned from the training data.
As expected, the number of dimensions kept after the dimensionality reduction is larger than in the other tasks. 
The need of a high degree of complexity is also denoted by the poor results obtained by using ridge regression in the readout training.
From Fig. \ref{fig:MSO_train}, we observe the convergence rate to be faster in models equipped with $\nu$-SVR, which obtain better results both in validation and in testing phase.
This symmetry on performances on test and validation reflects the scarce effectiveness of the regularization constraints for this task. 

% :::::::::::::::::::::::::::::::::::: TABLE RESULTS ::::::::::::::::::::::::::::::::::::
\bgroup
\def\arraystretch{.9} %vertical padding
\setlength\tabcolsep{.3em} %horizontal padding
\begin{center}
\begin{table}[ht]\scriptsize
    \centering
    \caption{\footnotesize Average prediction results obtained on the test set.
    The table contains the following fields: method for readout training (RT),
    dimensionality reduction procedure (DM),
    spectral radius of the reservoir ($\rho$),
    neurons in the reservoir ($N_r$),
    noise in ESN state update ($\xi$),
    scaling of input,
    teacher and feedback weights ($\omega_i$, $\omega_o$, $\omega_f$),
    dimensionality ($d$),
    kPCA kernel width ($\gamma_\text{k}$),
    $\mathrm{L}_2$ norm regularization factor ($\lambda$),
    $\nu$-SVR parameters ($C$, $\nu$, $\gamma_\text{r}$).
    Best results are highlighted in bold.}
    \vspace{0.2cm}
    \begin{tabular}{r|cccccccccccccc|p{1.05cm}p{1.5cm}}

    \cmidrule[1.5pt]{1-17}
      \textbf{}                           & %1
      \textbf{RT}                         & %2
      \textbf{DR}                         & %3
      $\mathbf{N_r}$                      & %4
      $\boldsymbol{\rho}$                 & %5
      $\boldsymbol{\xi}$                  & %6
      $\boldsymbol{\omega_i}$             & %7
      $\boldsymbol{\omega_o}$             & %8
      $\boldsymbol{\omega_f}$             & %9
      $\boldsymbol{\theta}^{\mathbf{(d)}}$& %10
      $\boldsymbol{\gamma}_\text{k}$      & %11
      $\boldsymbol{\lambda}$              & %12
      $\mathbf{C}$                        & %13
      $\boldsymbol{\nu}$                  & %14
      $\boldsymbol{\gamma}_\text{r}$      & %15
      \multicolumn{2}{l}{\textbf{Error}}  \\%16-17
    \cmidrule[1.5pt]{1-17}

    \multirow{6}{*}{\rotatebox[origin=c]{90}{MG}}
    %    2        &    3    &  4  &   5   &   6   &   7   &   8   &   9   & 10  &   11  &  12   &  13   &  14   &  15   &     16                  &       17
    & ridge reg.  & --      & 378 & 1.22  & 0.0   & 0.55  & 0.212 & 0.25  & --  & --    & 0.625 & --    & --    & --    & $3.064E^{-2}$           & $\pm 1.648E^{-4}$           \\
    & ridge reg.  & PCA     & 318 & 1.214 & 0.042 & 0.807 & 0.840 & 0.355 & 89  & --    & 0.294 & --    & --    & --    & $6.483E^{-3}$           & $\pm 9.126E^{-3}$           \\
    & ridge reg.  & $k$PCA  & 445 & 1.148 & 0.0   & 0.339 & 0.202 & 0.422 & 33  & 0.050 & 0.521 & --    & --    & --    & $4.283E^{-3}$           & $\pm 1.893E^{-4}$           \\
    & $\nu$-SVR   & --      & 490 & 1.051 & 0.0   & 0.873 & 0.588 & 0.568 & --  & --    & --    & 0.234 & 5.346 & 0.868 & $1.700E^{-1}$           & $\pm 5.171E^{-2}$           \\
    & $\nu$-SVR   & PCA     & 474 & 1.044 & 0.0   & 0.604 & 0.807 & 0.350 & 3   & --    & --    & 2.535 & 0.340 & 0.825 & $\mathbf{3.210E^{-5}}$  & $\pm \mathbf{3.122E^{-5}}$  \\
    & $\nu$-SVR   & $k$PCA  & 466 & 0.738 & 0.0   & 0.373 & 0.664 & 0.069 & 14  & 0.059 & --    & 7.975 & 0.448 & 0.299 & $4.902E^{-4}$           & $\pm 9.619E^{-6}$           \\
    \cmidrule[.5pt]{1-17}

    \multirow{6}{*}{\rotatebox[origin=c]{90}{NARMA}}
    %    2        &    3    &  4  &   5   &   6   &   7   &   8   &   9   & 10  &   11  &  12   &  13   &  14   &  15   &     16                  &       17
    & ridge reg.  & --      & 406 & 1.031 & 0.053 & 0.19  & 0.231 & 0.194 & --  & --    & 0.163 & --    & --    & --    & $3.759E^{-1}$           & $\pm 1.409E^{-3}$           \\
    & ridge reg.  & PCA     & 409 & 0.934 & 0.016 & 0.135 & 0.127 & 0.073 & 22  & --    & 0.963 & --    & --    & --    & $3.791E^{-1}$           & $\pm 4.887E^{-4}$           \\
    & ridge reg.  & $k$PCA  & 342 & 0.887 & 0.018 & 0.167 & 0.407 & 0.0   & 225 & 0.008 & 0.001 & --    & --    & --    & $1.024E^{-1}$           & $\pm 1.542E^{-3}$           \\
    & $\nu$-SVR   & --      & 440 & 0.928 & 0.015 & 0.129 & 0.603 & 0.031 & --  & --    & --    & 4.332 & 0.271 & 0.017 & $7.298E^{-2}$           & $\pm 7.901E^{-4}$           \\
    & $\nu$-SVR   & PCA     & 433 & 0.952 & 0.0   & 0.107 & 0.207 & 0.267 & 274 & --    & --    & 4.099 & 0.134 & 0.028 & $6.438E^{-2}$           & $\pm 6.254E^{-4}$           \\
    & $\nu$-SVR   & $k$PCA  & 460 & 0.962 & 0.002 & 0.1   & 0.302 & 0.037 & 163 & 0.028 & --    & 4.281 & 0.752 & 0.420 & $\mathbf{5.852E^{-2}}$  & $\pm \mathbf{1.475E^{-3}}$  \\
    \cmidrule[.5pt]{1-17}

    \multirow{6}{*}{\rotatebox[origin=c]{90}{MSO}}
    %    2        &    3    &  4  &   5   &   6   &   7   &   8   &   9   & 10  &   11  &  12   &  13   &  14   &  15   &     16                  &       17
    & ridge reg.  & --      & 298 & 1.148 & 0.008 & 0.345 & 0.147 & 0.045 & --  & --    & 0.438 & --    & --    & --    & $9.427E^{-1}$           & $\pm 1.675E^{-2}$           \\
    & ridge reg.  & PCA     & 499 & 1.141 & 0.017 & 0.187 & 0.184 & 0.309 & 438 & --    & 0.601 & --    & --    & --    & $7.642E^{-1}$           & $\pm 1.189E^{-1}$           \\
    & ridge reg.  & $k$PCA  & 454 & 1.053 & 0.003 & 0.137 & 0.169 & 0.184 & 407 & 0.040 & 0.117 & --    & --    & --    & $5.959E^{-1}$           & $\pm 3.233E^{-2}$           \\
    & $\nu$-SVR   & --      & 444 & 1.0   & 0.001 & 0.114 & 0.1   & 0.09  & --  & --    & --    & 3.714 & 0.282 & 0.037 & $\mathbf{2.353E^{-1}}$  & $\pm \mathbf{1.609E^{-2}}$  \\
    & $\nu$-SVR   & PCA     & 459 & 1.147 & 0.001 & 0.174 & 0.144 & 0.276 & 225 & --    & --    & 6.373 & 0.38  & 0.601 & $7.091E^{-1}$           & $\pm 2.182E^{-2}$           \\
    & $\nu$-SVR   & $k$PCA  & 480 & 1.027 & 0.032 & 0.1   & 0.627 & 0.011 & 135 & 0.002 & --    & 3.529 & 0.204 & 0.495 & $4.860E^{-1}$           & $\pm 2.082E^{-2}$           \\
    \cmidrule[1.5pt]{1-17}

    \end{tabular}
    \label{tab:results}
\end{table}
\end{center}
\egroup
% ::::::::::::::::::::::::::::::::::::::::::::::::::::::::::::::::::::::::::::::::::::::

%%%%%%%%%%%%%%%%%%%%%%%%%%%%%%%%%%%%%%%%%%%%%%%%%%%%%%%%%
%%%%%%%%%%%%%%%%%%%%% 5. DISCUSSION %%%%%%%%%%%%%%%%%%%%%
%%%%%%%%%%%%%%%%%%%%%%%%%%%%%%%%%%%%%%%%%%%%%%%%%%%%%%%%%
\section{Discussion}
\label{sec:discussion}

To understand the mechanics and the effectiveness of the proposed architecture, we analyze the results through the theory of nonlinear time-series analysis, which offer powerful methods to retrieve dynamical information from time-ordered data \cite{bradley2015nonlinear}.
The objective of time-series analysis is to reconstruct the full dynamics of a complex nonlinear dynamical system, starting from a measurement of only one of its variables.
In fact, in many cases it is possible to observe only a subset of the components necessary to determine the time-evolution law which governs the dynamical system.

The main idea which inspires this analysis is that a dynamic system is completely described by the time-dependent trajectory in its phase space.
Hence, a recurrent neural network that is capable of reconstructing with a high degree of accuracy the dynamic attractor can calculate future states assumed by the system, given a state at any particular moment.

A frequently used method for phase space reconstruction is the \emph{delay-coordinate embedding}, which provides an estimation of the attractor that is topologically identical to the true one.
From this reconstruction, it is possible to infer several properties of the hidden dynamical system, which are invariant under diffeomorhpism.
We refer to these measures as the \emph{dynamical invariants} of the system.
The most commonly studied are the \emph{fractal dimension} of the attractor, the \emph{Lyapuanov exponents} and the \emph{R\'{e}nyi entropy}.
In the following, we briefly introduce the delay-coordinate embedding procedure and two approaches used to estimate the aforementioned dynamical invariants.
We refer the interested reader to \cite{kantz2004nonlinear, gao2007multiscale} for a comprehensive overview of these methods and many other aspects of time-series analysis.

\paragraph{Delay-coordinate embedding:}
a dynamical system is characterized by a time-evolution law, which determines its trajectory in the phase space.
Each specific state of the system at time $t$ is defined by a $d$-dimensional vector in the state space: $\mathbf{s}(t) = [s_1(t), \dots, s_d(t)]^T$, being $d$ the number of variables of the system.
The delay-coordinate embedding method allows to reconstruct such state vectors from a time-discrete measurement of only one generic smooth function of the state space \cite{packard1980geometry}.
Given a discrete time-series $\x = \{x(i \Delta t)\}_{i=1}^N$ evenly sampled at rate $\Delta t$, the embedding is defined as:
\begin{equation}
    \label{eq:embedding}
    \hat{s}(i) = \sum \limits_{j=1}^m x(i + (j-1)\tau_e) \mathbf{e}_j,
\end{equation}
where $m$ is the embedding dimension, $\tau_e$ is the time delay and $\mathbf{e}_j$ form an orthonormal basis in $\mathbb{R}^m$.

With a proper choice of embedding parameters $m$ and $\tau_e$, Taken theorem guarantees the existence of a diffeomorhpism between the real and reconstructed dynamic \cite{takens1981detecting}.
A sufficient condition for a correct reconstruction is $m \geq 2d +1$.
The value of $m$ is usually computed with the false nearest-neighbors algorithm \cite{rhodes1997false}, which provides an estimation of the smallest sufficient embedding dimension.
On the other hand, a suitable time-delay $\tau_e$ can be estimated looking at the first zero of the autocorrelation function of $\x$ or by relying on nonlinear time dependencies, such as the mutual information \cite{cao1997practical}.

\paragraph{Correlation dimension:}
dimension is a measurement invariant under diffeomorhpism that allows to quantify the similarity of geometrical objects. 
Attractors of dissipative chaotic systems often exhibit complicated geometries (hence the name \emph{strange}) which are contained in a fractal dimension $D_q$, called R\'{e}nyi dimension \cite{renyi1959dimension}.
An efficient estimator of fractal dimensions is Grassberger-Procaccia algorithm \cite{grassberger2004measuring}, which computes the correlation dimension $D_2$ through the correlation sum $C_2$:
\begin{equation}
 \label{eq:corr_sum}
    C_2(m,\epsilon) = \frac{1}{2N_\epsilon(N_\epsilon-\tau_c)} \sum \limits_{i} \sum \limits_{j < i-\tau_c} \boldsymbol{\Theta} \left( \epsilon - \| x(i) - x(j) \| \right).
\end{equation}
The temporal spacing parameter $\tau_c$ it chosen to ensure temporal independence between samples, $\boldsymbol{\Theta}$ is the Heaviside function and $\epsilon$ is the dimension of a set of $N_\epsilon$ small boxes used to cover the geometric shape of the attractor. If $m \geq D_2$, $C_2(m,\epsilon) \propto \epsilon^{D_2}$.

\paragraph{Lyapuanov exponent:}
the Lyapuanov spectrum $\{ \lambda_1, \dots, \lambda_d \}$ is another invariant measure that characterizes the predictability of a dynamical system.
Lyapuanov exponents quantify the rate of separability of two infinitesimal close trajectories and are closely related to the 2\textsuperscript{nd} order R\'{e}nyi entropy $K_2$: $K_2 \leq \sum_{\lambda_i > 0} \lambda_i$.
This quantity measures the number of possible trajectories that the system can take for a given number of time steps in the future. 
A perfectly deterministic can only evolve along one possible trajectory and hence $K_2 = 0$.
In contrast, for purely stochastic systems the number of possible future trajectories increases to infinity, so $K_2 \rightarrow \infty$.
Chaotic systems are characterized by a finite value of $K_2$, as the number of possible trajectories diverges but not as fast as in the stochastic case.

The largest Lyapuanov exponent (LLE) $\lambda_1$ is a good estimate of $K_2$, and its sign determines whether a system is chaotic or not.
The so-called \emph{direct methods} can be used to compute $\lambda_1$ by estimating the divergent motion of the reconstructed space, without fitting a model to the data \cite{wolf1985determining, parlitz1998nonlinear}.
In particular, the average exponential growth of the distance of neighboring orbits can be studied on a logarithmic scale by monitoring the prediction error $p(t)$:
\begin{equation}
    \label{eq:pred_error}
    p(t) = \frac{1}{N} \sum \limits_{k=1}^N \log_2 \left( \frac{\| \x[k+t] - \x_\text{nn}^\text{[k]}[t] \|}{\| \x[k] - \x_\text{nn}^\text{[k]} \|} \right),
\end{equation}
being $\x_\text{nn}^\text{[k]}$ the nearest neighbor of $\x$ at time $k$. The LLE is estimated as $\lambda_1 \propto p(t)/T$ with $t \in [1,T]$, where $T$ is the forecast horizon within which the divergence of the trajectory in the phase space is evaluated.

% -------------- ESN phase space reconstruction --------------
\subsection{ESN phase space reconstruction}

In the following, we analyze two chaotic time-series generated by the Lorenz and the Moore–Spiegel system respectively.
We evaluate the accuracy of the phase space reconstruction performed with our ESN by comparing the topological properties of the true attractor of the dynamic, with the one obtained by applying a dimensionality reduction to the network reservoir.
The equivalence of attractors geometries are computed by measuring the dynamical invariants, estimated through the correlation sum and the divergent motion of the reconstructed spaces. 

In the following, we refer to \emph{true attractor}, as the trajectory in the phase space generated directly by the differential equations of the dynamic system.
With \emph{delay-embedding attractor} we refer at the trajectory described by the embedding, generated with the delay-coordinate procedure.
Finally, \emph{ESN attractor} is the trajectory spanned by the component of the multivariate vector $\bar{\mathbf{h}}$.
The latter is the output of the dimensionality reduction procedure applied to the multivariate vector $\mathbf{h}$, which contains the sequence of the states of the reservoir (see Sect. \ref{sec:architecture}).
For these tests we considered only the component of the loss function relative to the prediction error, by setting $\alpha = 0$ in Eq.~\ref{eq:loss}, and we fixed the number of dimensions in PCA and kPCA to 3.
Finally, to further empathize the effectiveness of the architecture proposed, we also consider the phase space reconstruction obtained directly from $\mathbf{h}$, in the case where the reservoir contains only 3 neurons ($N_r = 3$). 

%::::::::::::::: FIG LORENTZ ::::::::::::::::
\begin{figure}[ht]
\centering
	\subfigure[True attractor trajectory.]{
	\includegraphics[width=0.3\textwidth]{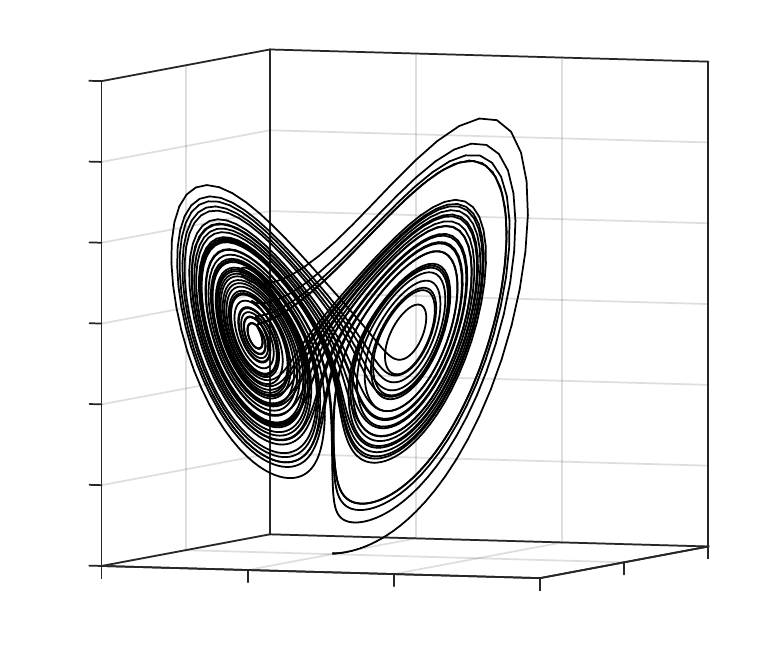}
	\label{fig:lorentz_true}
	}
	~
	\subfigure[Time-delay embedding trajectory.]{
	\includegraphics[width=0.3\textwidth]{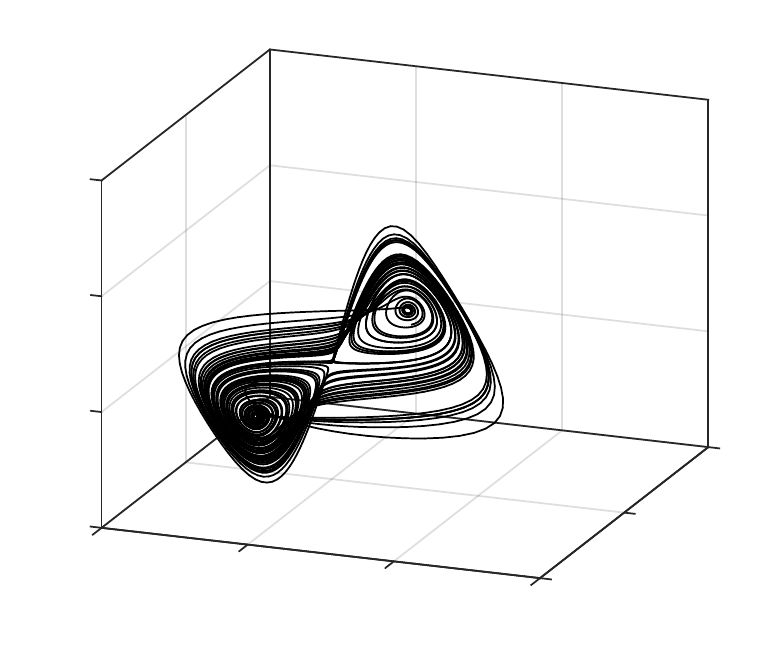}
	\label{fig:lorentz_emb}
	}
	
	\subfigure[ESN+PCA trajectory.]{
	\includegraphics[width=0.3\textwidth]{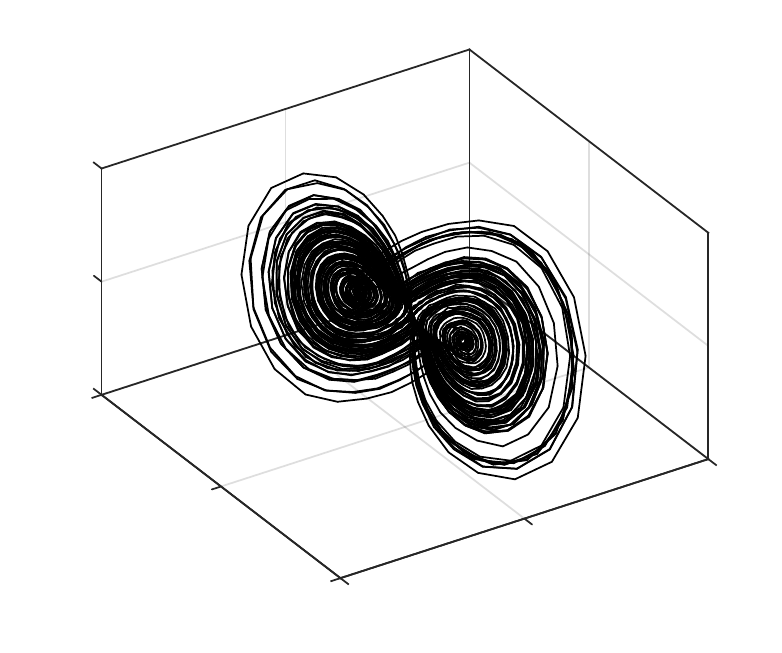}
	\label{fig:lorentz_esn}
	}
	~
	\subfigure[ESN with 3 neurons trajectory.]{
	\includegraphics[width=0.3\textwidth]{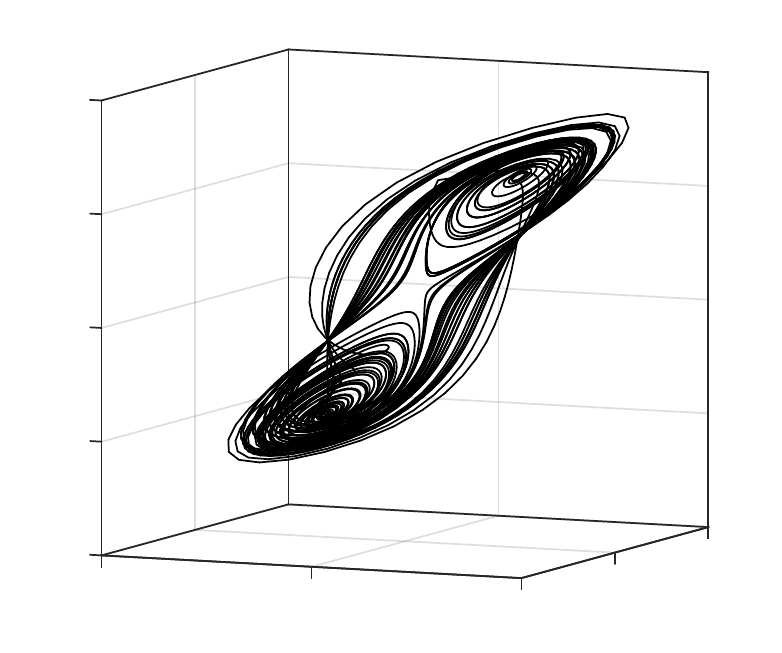}
	\label{fig:lorentz_small}
	}
				
\caption{\footnotesize trajectory of the attractors of the Lorenz dynamical system in the phase space. In (a), the true trajectory, which is computed directly from the ordinary differential equations of the system. In (b), the trajectory reconstructed using time-delay embedding. In (c), the trajectory generated by the internal state of ESN internal state, on the subspace defined by the first 3 components of the PCA. In (d), the trajectory described by the internal state of an ESN with a small reservoir with 3 neurons.}
\label{fig:lorentz}
\end{figure}
%::::::::::::::::::::::::::::::::::::::::::::

% Lorentz
\paragraph{Lorenz:} the system is governed by the following ordinary differential equations:
%%:::::::::::::: TAB INVARIANTS ::::::::::::::
\bgroup
\def\arraystretch{1.1} %vertical padding
\setlength\tabcolsep{.4em} %horizontal padding
\begin{center}
\begin{table}[tbp]\scriptsize
	\centering
	\caption{\footnotesize Correlation dimension ($D_2$) and largest Lyapuanov exponent (LLE) of the attractors of Lorenz and Moore-Spiegel dynamical systems. Each invariant is estimated on the trajectories generated by: the ordinary differential equations (True); the dime-delay embedding (Emb); the ESN reservoir state, whose dimensionality is reduced using PCA (ESN+PCA) or $k$-PCA (ESN+$k$PCA); the internal state of an ESN with a small reservoir with 3 neurons (ESN small).}
	\vspace{0.2cm}
	\begin{tabular}{r|cccccc}
	\cmidrule[1.5pt]{1-7}
	\textbf{System} & \textbf{Invariant} & \textbf{True} & \textbf{Emb} & \textbf{ESN+PCA} & \textbf{ESN+\emph{k}PCA} & \textbf{ESN small} \\
	\cmidrule[1.5pt]{1-7}
	\multirow{2}{*}{Lorenz} & $D_2$ & $2.068 \pm 4e^{-6}$ & $1.8871 \pm 8e^{-6}$ & $2.1722 \pm 3e^{-6}$ & $ 1.8614\pm 5e^{-6}$ & $1.6044 \pm 1e^{-6}$ \\ 
	& LLE & $0.9056 \pm 5e^{-4}$ & $0.9181 \pm 6e^{-4}$ & $1.0397 \pm 5e^{-4}$ & $0.91496 \pm 8e^{-4}$ & $0.76138 \pm 3e^{-5}$ \\ 
	\cmidrule[.5pt]{1-7}
	\multirow{2}{*}{Moore-Spiegel} & $D_2$ & $1.9802 \pm 1e^{-6}$ & $0.83499 \pm 4e^{-6}$ & $0.87619 \pm 3e^{-6}$ & $0.95588 \pm 1e^{-6}$ & $0.63507 \pm 2e^{-7}$  \\ 
	& LLE & $0.00708 \pm 7e^{-4}$ & $0.7003 \pm 4e^{-4}$ & $0.51611 \pm 4e^{-4}$ & $0.54784 \pm 4e^{-4}$ & $0.75421 \pm 2e^{-5}$  \\ 
	\cmidrule[1.5pt]{1-7}
	\end{tabular}
	\label{tab:invariants}
\end{table}
\end{center}
\egroup
\begin{equation}
\label{eq:lorentz}
\frac{\text{d} x}{\text{d}t} = \sigma (y -x), \; \frac{\text{d} y}{\text{d}t} = x (\rho -z) - y, \; \frac{\text{d} z}{\text{d}t} = xy - \beta z,
\end{equation}
    where variables $x$, $y$ and $z$ define the state of the system, while $\sigma$, $\rho$ and $\beta$ are system parameters.
In this work we set  $\sigma = 10$ , $\beta = 8/3$  and $\rho = 28$, values for which the system exhibits chaotic behavior.

%::::::::::::::::::::::::::::::::::::::::::::

% Lorentz discussion
Fig. \ref{fig:lorentz} depicts the geometric shapes of the true attractor, the delay-embedding attractor, the two ESN attractors, generated using a dimensionality reduction or a reservoir with 3 neurons.
As is it possible to observe visually, both the embedding and ESN with dimensionality reduction manage to reconstruct well the trajectory described by the differential equations of the dynamic system.
To quantify formally this similarity, we compute on the dynamical invariants previously introduced each attractor.
In Tab.~\ref{tab:invariants}, we report for each phase space trajectory the estimated correlation dimension and the largest Lyapuanov exponent, which as previously discussed, represents a good approximation of the $K_2$ entropy.
Due to the stochastic nature of the approaches adopted for estimating these quantities, we repeated the procedure 10 different times and we report their average values and the standard deviations.
As we can see from the results, both the trajectories described by $\bar{\mathbf{h}}$ in the subspace computed using PCA and kPCA generate an attractor whose dynamic invariants are well approximated.
In particular, the accuracy of the reconstruction is comparable to the one obtained by the classic time-delay embedding method and in some case it is even better.
The standard deviations in the measurements of both correlation dimension and LLE are very small, which indicates a high degree of reliability on both measurements.
For what concerns the ESN with 3 neurons, the trajectory described is more ``flat'', as it can be seen in the figure.
This is confirmed by the estimated correlation dimension and LLE, whose values are much lower than in the other cases.
This denotes that the reconstructed dynamic is not rich enough, a symptom that the complexity and the memory of the network is not sufficient to model the underlying system.

%:::::::::::::: FIG MS ::::::::::::::
\begin{figure}[ht]
\centering
	\subfigure[True attractor trajectory.]{
	\includegraphics[width=0.3\textwidth]{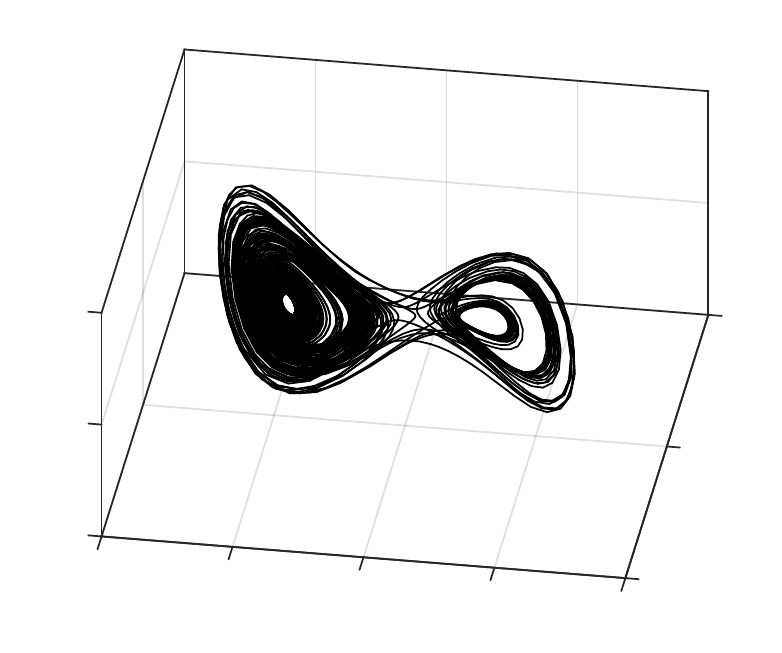}
	\label{fig:ms_true}
	}
	~
	\subfigure[Time-delay embedding trajectory.]{
	\includegraphics[width=0.3\textwidth]{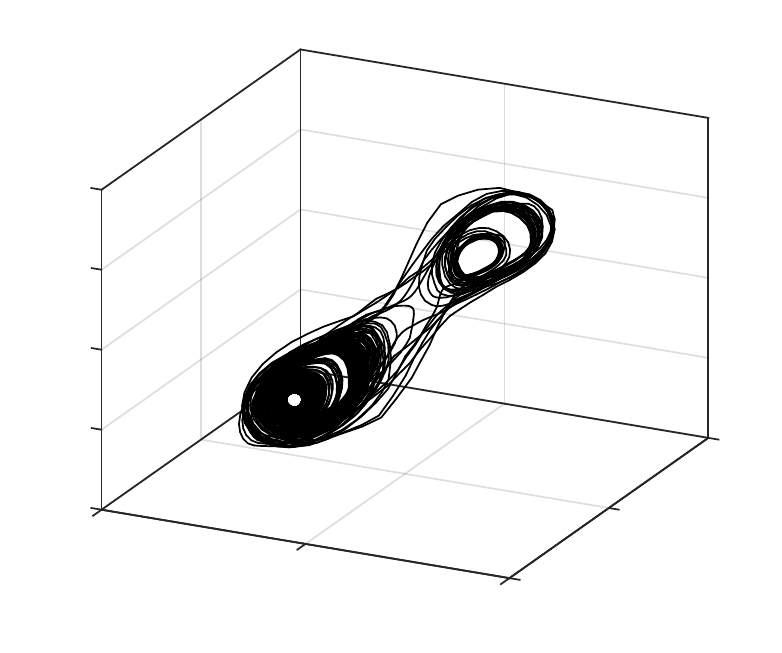}
	\label{fig:ms_emb}
	}
	
	\subfigure[ESN+PCA trajectory.]{
	\includegraphics[width=0.3\textwidth]{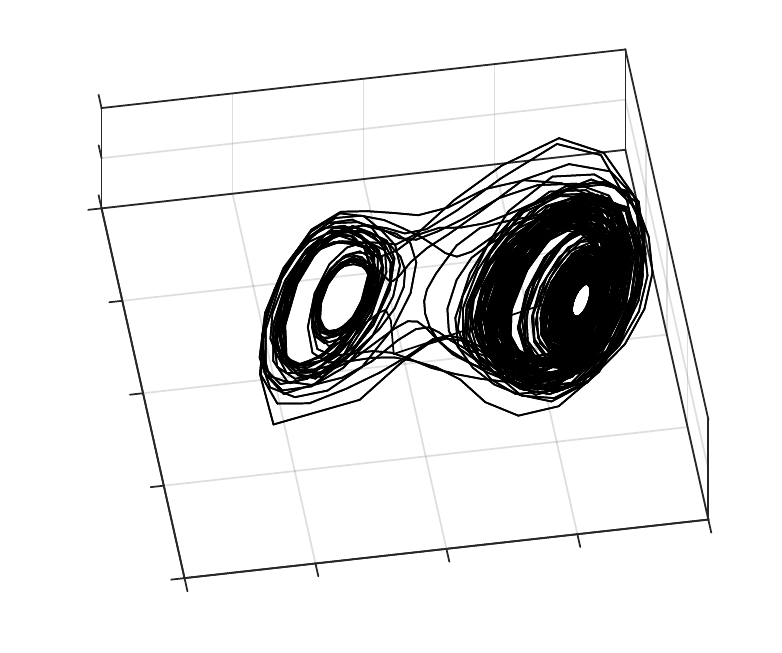}
	\label{fig:ms_esn}
	}
	~
	\subfigure[ESN with 3 neurons trajectory.]{
	\includegraphics[width=0.3\textwidth]{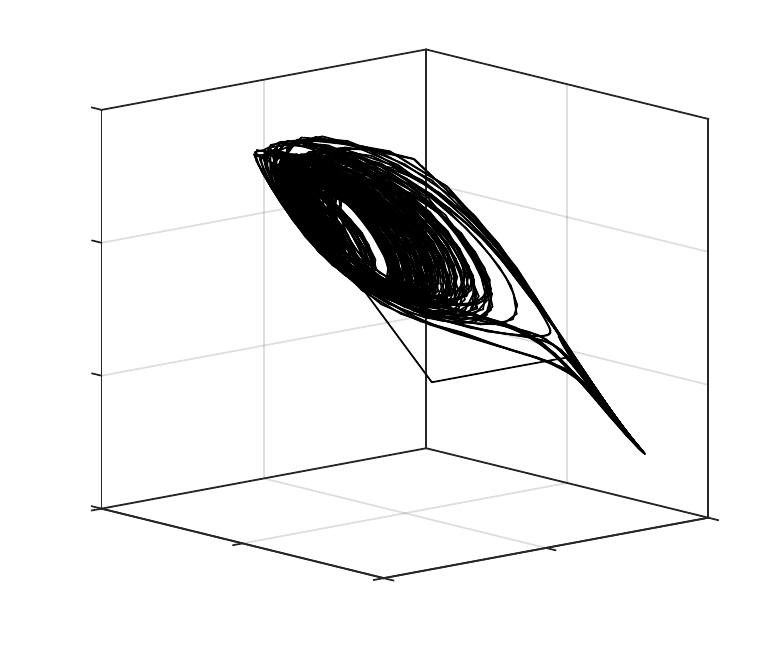}
	\label{fig:ms_small}
	}			
\caption{\footnotesize trajectory of the attractors of the Moore-Spiegel dynamical system in the phase space. In (a), the true trajectory, which is computed directly from the ordinary differential equations of the system. In (b), the trajectory reconstructed using time-delay embedding. In (c), the trajectory generated by the internal state of ESN internal state, on the subspace defined by the first 3 components of the PCA. In (d), the trajectory described by the internal state of an ESN with a small reservoir with 3 neurons.}
\label{fig:MS}
\end{figure}
%:::::::::::::::::::::::::::::::::::::

% Moore–Spiegel
\paragraph{Moore–Spiegel:} this dynamical systems manifests interesting synchronization properties, generated by complicated patterns of period-doubling, saddle-node and homoclinic bifurcations \cite{balmforth1997synchronizing}.
The differential equations which governs system dynamics are the following:
\begin{equation}
\label{eq:moorespiegel}
\frac{\text{d} x}{\text{d}t} = y, \; \frac{\text{d} y}{\text{d}t} = z, \; \frac{\text{d} z}{\text{d}t} = -z -(t-r+rx^2)y -tx,
\end{equation}
where $x$, $y$ and $z$ form the state of the system and $r$ and $t$ are the parameters of the model. 
In this study, we set $r = 100$, $b = 10$ and $c = 14$, for which the dynamics of the system exhibits a chaotic behavior.

% Moore–Spiegel discussion
In Fig. \ref{fig:MS} we show the shape of the attractors of the dynamic, evaluated directly on the differential equations of the system, on the time-delay embedding, on the internal state of the ESN reduced through PCA and on the state of the ESN with 3 neurons.
In this second test, the reconstructed trajectories of the Moore–Spiegel system are more jagged and irregular, with respect to the original one.
This suggest a poorer approximation of the true dynamic of the system and is confirmed by the results in Tab.~\ref{tab:invariants}.
Compared to the Lorenz case, the dynamical invariants estimated on the time-delay embedding and on ESN state trajectories approximate with less accuracy the real ones.
The reconstructed attractors have a lower correlation dimension, which usually denotes a poor embedding \cite{marwan2007recurrence}.
However, it is worth to notice that the two attractors reconstructed by the ESN+PCA and ESN+kPCA have a larger $C_2$ value than the time-delay embedding and hence they approximate better the true dynamics.
For what concerns the LLE, the estimated value in each reconstructed dynamic is larger than in the original one.
This means that both the time-delay embedding and the ESNs generate a more chaotic dynamic, as is also reflected by the jagged trajectories in Fig. \ref{fig:MS}.
Even in this case, however, LLE is better approximated by ESN+PCA and ESN+kPCA than by the time-delay embedding.
Like before, the standard deviations of the estimates of the two dynamical invariants is very small, which provides a high degree of confidence on the measurements.
For what concerns the trajectory described by the ESN state with a small reservoir of 3 neurons, the geometric properties of the reconstruct attractor are even more different from the real ones.
This confirm that also in this case such a small amount of neurons cannot catch the dynamic properties of the system to be modeled.

% concluding remarks
As a concluding remark, it is important to understand another aspect of the utility of the ESN in reproducing the attractor of the system dynamic.
In fact, this provides a valid alternative to the standard approach based on the time-delay embedding for reconstructing the phase of the system, which presents several caveats and pitfalls \cite{bradley2015nonlinear}.
This a fundamental tool for a wide set of applications, where an accurate estimation of the phase space of the system is required \cite{kantz2004nonlinear}.

%%%%%%%%%%%%%%%%%%%%%%%%%%%%%%%%%%%%%%%%%%%%%%%%%%%%%%%%%%
%%%%%%%%%%%%%%%%%%%%% 6. CONCLUSIONS %%%%%%%%%%%%%%%%%%%%%
%%%%%%%%%%%%%%%%%%%%%%%%%%%%%%%%%%%%%%%%%%%%%%%%%%%%%%%%%%
\section{Conclusions and future directions}
\label{sec:conclusion}

In this work we have presented a new framework for training an Echo State Network,
  which enhances its generalization capabilities through the regularization
  constraints introduced by the smoothing effect of a dimensionality reduction
  procedure.
Through a series of test on benchmark dataset, we have demonstrated how the
  proposed architecture can achieve better prediction performance in different
  contexts.
Successively, we provided a theoretically grounded explanation of the functioning
  of the proposed architecture, based on the theory of nonlinear time-series analysis.
By studying the dynamical properties of the network under this novel perspective,
  we showed that through an ESN it is possible to reconstruct the phase space of
  the dynamic system; this offers a solid, yet simple alternative to the time-delay
  embedding procedure.

We believe that this work could be useful not only to enhance the prediction capabilities
  of an ESN, but also provide a new instrument for analysis of dynamical systems.
% Possible future directions include trying more advanced methods for dimensionality   reduction and manifold learning, in order to improve the results obtained both   in terms of prediction accuracy and phase space reconstruction. - repetition
As a follow-up of a recent work focused on identifying the edge of criticality of
  an ESN by evaluating the Fisher information on the state matrix \cite{livi2016determination},
  we plan to study the criticality using more reliable Fisher Information Matrix
  estimators, which are capable of working only on space with few dimensions
  (e.g., \cite{har2016nonparametric}).
We also plan on investigating other dimensionality reduction methods, manifold learning
  and semi-supervised learning approaches to shrink and regularize the output of the network recurrent layer \cite{belkin2006manifold,bengio2004out}.
Finally, as a future work, we propose to use different dimensionality reduction techniques in parallel and combine their result through a single reservoir to produce the final result.

\bibliographystyle{abbrvnat}
\bibliography{Biblio}
\end{document}